# ZTree: a subgroup identification based decision tree learning framework

Eric Cheng (New York University, Email: erc7558@nyu.edu)
and
Jie Cheng (Statistical and Quantitative Sciences, Takeda Pharmaceuticals)

## Abstract

Decision trees are a commonly used class of machine learning models valued for their interpretability and versatility, capable of both classification and regression. We propose ZTree, a novel decision tree learning framework that replaces CART's traditional purity based splitting with statistically principled subgroup identification. At each node, ZTree applies hypothesis testing (e.g., z-tests, t-tests, Mann-Whitney U, log-rank) to assess whether a candidate subgroup differs meaningfully from the complement. To adjust for the complication of multiple testing, we employ a cross-validation-based approach to determine if further node splitting is needed. This robust stopping criterion eliminates the need for post-pruning and makes the test threshold (z-threshold) the only parameter for controlling tree complexity. Because of the simplicity of the tree growing procedure, once a detailed tree is learned using the most lenient z-threshold, all simpler trees can be derived by simply removing nodes that do not meet the larger z-thresholds. This makes parameter tuning intuitive and efficient. Furthermore, this z-threshold is essentially a p-value, allowing users to easily plug in appropriate statistical tests into our framework without adjusting the range of parameter search. Empirical evaluation on five large-scale UCI datasets demonstrates that ZTree consistently delivers strong performance, especially at low data regimes. Compared to CART, ZTree also tends to grow simpler trees without sacrificing performance. ZTree introduces a statistically grounded alternative to traditional decision tree splitting by leveraging hypothesis testing and a cross-validation approach to multiple testing correction, resulting in an efficient and flexible framework.

## Introduction

Decision trees are a commonly used class of machine learning models valued for their interpretability and versatility. Unlike many modern machine learning models that are often regarded as "black boxes," decision trees produce rule-based structures that are simple to understand and communicate. One can easily trace through the tree structure to see how the algorithm yields a certain prediction. A key strength of decision trees is their natural ability to handle both categorical and continuous data, as well as both classification and regression tasks.

Most traditional and popular decision tree algorithms, such as CART (Classification And Regression Tree; Breiman et al. 1984), are based on measures of node purity, using criteria like

Gini impurity or information gain to determine splits. Not all decision tree frameworks rely on these metrics however. In this work, we propose a novel framework for decision tree construction grounded in subgroup identification. Rather than optimizing for purity, our method focuses on identifying statistically meaningful subgroup effects within the data.

To assess the relevance of a potential split, we apply statistical hypothesis testing. For example, for categorical outcomes, we can utilize two-proportion z-tests; for continuous outcomes, we apply either two sample t-tests or the Mann-Whitney U tests, depending on distributional assumptions. This statistical testing framework provides a principled way to evaluate the meaningfulness of each split beyond heuristic criteria.

A central challenge of this approach is how to adjust for multiple testing effectively when deciding if a node should be further splitted or not. Many subgroup hypotheses need to be assessed and these hypotheses are often correlated (e.g., subgroup criteria age>65, age>70, and high_blood_pressure="yes"). As a result, the standard Bonferroni correction is almost always too conservative for such an application. To address this, we introduce a cross-validation-based strategy that adjusts for multiple comparisons effectively regardless of the dependencies among subgroup hypotheses. This adjustment ensures that detected subgroup effects have certain statistical meaning and not artifacts due to multiple hypothesis testing.

## Overview and related work

The proposed algorithm builds upon the classical decision tree framework of CART, by employing a greedy, recursive partitioning strategy to grow binary trees for both classification and regression tasks. However, unlike CART, which relies on node impurity measures (e.g., Gini index or entropy) to guide splitting decisions, our method uses statistical hypothesis testing to evaluate subgroup effect (difference between subgroup and the rest of the instances) and determine both whether a node should be split, and how to find the best subgroup to split the data. Another key distinction is that our algorithm does not employ post-pruning; instead, model complexity is regulated during tree growth through statistical subgroup effect testing.

There is a well-established body of research exploring the use of statistical tests to guide node splitting in decision trees. Notable examples include CHAID (Kass, 1980), GUIDE (Loh, 2002), QUEST (Loh and Shih, 1997), CTree (Hothorn et al., 2006), SIDES (Lipkovich et al, 2011), QUINT (Dusseldorp and Van Mechelen 2014), and MOB (Zeileis et al., 2008). These algorithms often forego post-pruning, as overfitting is mitigated during the tree-building process itself—typically through the use of techniques such as permutation testing or bootstrapping to assess split validity. This design allows for flexible integration of specialized tests, such as those tailored to detecting differential treatment effects, as demonstrated in GUIDE. Some methods, such as SIDES and QUINT, are explicitly developed for subgroup identification in the context of treatment effect heterogeneity. A comprehensive review of subgroup identification techniques can be found in (Lipkovich and Dmitrienko, 2017).

Cross-validation remains a standard approach in supervised learning for both model evaluation and hyperparameter tuning. However, cross-validation is used in a slightly different way in the proposed method – we incorporate cross-validation to obtain unbiased ("un-inflated") test statistics for the subgroup effect. Additionally, another layer of cross-validation can be used to tune the only hyperparameter in our model, which is the threshold for accepting the null hypothesis (i.e., stop splitting since no strong enough subgroup effect), or rejecting the null hypothesis (i.e., use the identified subgroup to split). This threshold directly influences model complexity, allowing the user to control the granularity of the resulting decision tree. To distinguish the two cross-validation procedures, we refer to the former one as internal cross-validation and the latter one as external cross-validation.

# The proposed decision tree learning framework

The proposed DT learning framework is based on CART but without any post-pruning. The main difference lies in the splitting criteria and stopping rule. Instead of using node purity based splitting criteria, we use statistical tests to evaluate if a subgroup of instances is different from the rest of the instances, in terms of a predefined target (e.g., the mean of the y variable.) To adjust for multiple testing, a cross-validation procedure is applied for each node to get the cross-validated statistical test score. If the score is equal or larger than the predefined significance threshold, the instances of the node will be partitioned into two child-nodes, the identified subgroup and the rest. Otherwise, the current node becomes a leaf node.

The pseudo-code of the proposed DT learning framework is as follows.

*learnTree (data, statistical_test, threshold, search_depth) {*
*/\**
- *data: the input training data. In the format <x, y> or <x, y, trt>.*
    - *x: features;*
    - *y: target/outcome variable;*
    - *trt: binary treatment variable. Used for finding differential treatment effect subgroups*
- *statistical_test: the predefined statistical test to check if a subgroup is different from the rest of the instances*
- *threshold: the significance threshold for the statistical test*
- *search_depth: 1, 2, or 3. Default=1*
    - *1: univariate search (e.g., age>65).*

```
                ○   2: up to 2 variable combinations (sex=M & age>65).
                ○   3: up to 3 variable combinations.
*/
        node = new node;
        cv_test_score = cross-validate (data, statistical_test, search_depth);
        If (cv_test_score >= threshold) {
                model = Train (data, statistical_test, search_depth);
                <subgroup, rest of instances> = apply_model(model, data);
                node.left = learnTree(subgroup, statistical_test, threshold, search_depth);
                node.right = learnTree(rest_of_instances, statistical_test, threshold,
search_depth) ;
        }
        return node;
}

cross-validate (data, statistical_test, search_depth) {
        For each iteration of CV {
                <train_data, validation_data> = assign_train_validation(data);
                Model m = train (train_data);
                apply_model(m, validation_data); //assign subgroup membership

        }
        CV_test_score = statistical test( all instances where inSubgroup==True, all instances
where inSubgroup==false);
        Return (CV_test_score);
}

train (data, statistical_test, search_depth) {
        Find the best subgroup with the largest test score;
        Model m = the best subgroup criterion; // for example, age>65
        return m;
}

apply_model (model, data) {
        For each instance ins in data {
                If ins meets criterion of model
                        ins.inSubgroup = True;
                Else
                        ins.inSubgroup = False;
        }
        return <subgroup, rest_of_instances>;
}
```

# Statistical Tests

A notable advantage of the proposed decision tree (DT) learning framework is its flexibility in accommodating a variety of statistical tests. This modularity allows the algorithm to be easily adapted to different types of outcomes and modeling goals.

**1. Dataset Format: ⟨Features, Outcome⟩**

For the standard supervised learning tasks, the goal is to partition the instances into outcome subgroups. The choice of statistical test depends on the nature of the outcome variable:

- **Binary outcomes**: We employ the **two-proportion *z*-test** to assess whether the distribution of class labels differs significantly between candidate child nodes.

- **Continuous outcomes**: Both the **two-sample *t*-test** and the **Mann-Whitney *U* test** (a non-parametric alternative) are implemented to detect differences in mean or distribution of outcomes across splits.

- **Time-to-event outcomes**: The **log-rank test** is used to compare survival distributions between groups formed by a candidate split.

**2. Dataset Format: ⟨Feature, Outcome, Treatment⟩**

In scenarios involving treatment assignment, where the objective is to identify subgroups with heterogeneous treatment effects, we incorporate specialized statistical tests appropriate for binary, continuous, and time-to-event outcomes. These tests are based on the following general formula:

$$Z = \frac{\text{Treatment Effect}_{\text{Subgroup}} - \text{Treatment Effect}_{\text{Complement}}}{\text{Standard Error of the Difference}}$$

The treatment effect for subgroup or complement is:

$$\text{Treatment Effect} = \text{E}[Y \mid T = 1] - \text{E}[Y \mid T = 0]$$

Where Y is observed outcome and $T \in \{0,1\}$: treatment assignment (1 = treatment, 0 = control)

Standard Error of the Difference: $\text{SE}_{\text{diff}} = \sqrt{\text{Var}_{\text{Subgroup}} + \text{Var}_{\text{Complement}}}$

To facilitate a unified decision criterion across various types of statistical tests, we standardize all test statistics to the *z*-score scale. Although this conversion is not strictly necessary, it simplifies the tuning and comparison of split decisions across nodes and outcome types. For consistency, we refer to all statistical test values as *z*-scores throughout the remainder of this paper.

**Finding Cutoff Values for Continuous Features**

In the classic CART algorithm, all possible threshold values of a continuous variable are considered when evaluating potential splits. While this exhaustive approach can identify highly specific split points, it is computationally intensive and prone to overfitting, particularly in high-dimensional datasets or when sample sizes are small.

To address these challenges, our framework restricts the search space for continuous feature cutoffs. Specifically, we limit the number of candidate thresholds to a maximum of 20, selected approximately at every 5th percentile of the feature's empirical distribution. This percentile-based discretization strikes a balance between computational efficiency and model flexibility, allowing the algorithm to capture meaningful variation without overfitting to noise.

**Learning Trees of Varying Complexity**

The complexity of the decision trees produced by our framework can be efficiently controlled by adjusting the threshold applied to the cross-validated test scores during the split-versus-stop decision process. A lower threshold results in deeper, more complex trees, while a higher threshold yields simpler, more parsimonious models.

A key advantage of our approach is that, due to its simplicity and the absence of post-pruning, trees of multiple complexity levels can be derived simultaneously with minimal additional computational cost. Specifically, we first train a base decision tree using the smallest threshold of interest (e.g., 0.2), recording the cross-validated test score at each internal node. To obtain trees corresponding to higher thresholds (e.g., 0.3, 0.4, etc.), we simply traverse the base tree and prune nodes whose test scores fall below the desired threshold.

This strategy eliminates the need to retrain the model for each threshold value and enables efficient parameter tuning and model selection across a range of complexity levels.

For the internal CV, we apply 5 fold CV 10 times and calculate the mean CV performance. For the external CV, we apply a 10 fold CV once.

## Implementation of the proposed method

The proposed method was implemented in Java with a Python interface.

# Experiment

**Datasets**

To evaluate the prediction performance and other properties of the proposed method, we conducted an empirical comparison with the Scikit Learn CART decision tree learning algorithm 1.6.0 and the support vector machines algorithm (SVM), using 5 popular datasets from the UCI Machine Learning Repository (Kelly et al.).

The datasets are listed in Table 1. These datasets were chosen for several reasons: relatively large sample size (>10K instances), popularity, representing different application fields, and covering both classification and regression tasks. "Bike sharing" and "news popularity" are the two regression datasets. The target continuous variable of neither dataset conforms to normal distribution. Because of the extreme skewness of the "news popularity" dataset, we log transformed the target variable. The target variable "cnt" of the "bike sharing" is unchanged. For CART and SVM, nominal features were one-hot encoded and ordinal features were label-encoded. The ZTree implementation can handle categorical features out of the box.

| Dataset | Instances | Features | Continuous Features | Categorical Features | Task |
|---|---|---|---|---|---|
| Adult Income | 48842 | 12 | 4 | 8 | Classification |
| Bank Marketing | 41188 | 20 | 11 | 9 | Classification |
| Online Shopping | 12330 | 17 | 14 | 3 | Classification |
| Bike Sharing | 17389 | 12 | 4 | 8 | Regression |
| Online News Popularity (log transformed) | 39797 | 58 | 58 | 0 | Regression |

**Experiment setup**

We evaluated model performance using area under the ROC curve (AUROC) for classification tasks and root mean squared error (RMSE) for regression tasks, providing a standard basis for comparison with existing machine learning methods.

For each dataset, we randomly sampled 100 independent training sets of sizes 100, 300, 1000, and 3000 instances. The remaining data in each case was used as the independent test set, ensuring consistent evaluation across varying sample sizes. Note that the exact same training sets and testing sets were used for all three methods.

For each training set, 10 fold cross-validation was performed for parameter tuning in all three methods. The best parameters were then applied to train the optimal model using the whole training set. The optimal model is applied to the independent test set to get model performance.

For ZTree, the optimal z-threshold from the set [0.2, 0.4, 0.6,...3.0] was selected based on the cross-validated model performance. Z-threshold is the parameter that determines split or stop for each node, which is the only parameter that needs to be tuned.

For CART, hyperparameters of max_depth and min_samples_split were optimized with a 10-fold Bayes Cross Validation. To determine appropriate values for the hyperparameter tuning used in model training, we employed an empirical tuning strategy. A range of candidate depth values was explored based on prior experience, domain knowledge, and iterative experimentation.

For SVM hyperparameters, using a BayesCV search, the C value was optimized from the range 1e-3 to 1e3, and the gamma value was optimized from the range 1e-4 and 1. The C values cover 6 orders of magnitude, capturing a wide spectrum of regularization behaviors. The gamma value covers 4 orders of magnitude, searching both broad and tight decision boundaries.

# Results

Table 1 summarizes the mean performance of each method for each dataset/sample size setting. The best performance of each setting is shown in bold fonts. Figures 1 to 5 show the result of the 5 datasets, with each figure summarizing the result of one dataset. The top panel of each figure shows the mean performance (AUROC or RMSE) of 100 test sets for each method at each sample size. The error bars represent 95% confidence intervals of the mean. The bottom panel shows the mean decision tree depth of the trained models for CART-sklearn and ZTree. Again, the error bars represent 95% confidence intervals of the mean.

Figure 1 shows that for the Adult income dataset, SVM has the best performance for sample sizes 300, 1000, and 3000. ZTree and CART have similar performance at sample sizes 300, 1000 and 3000. ZTree has the best performance for sample size 100.

Figure 2 shows that for the bank marketing dataset, ZTree and SVM perform better than CART. ZTree may have an advantage for sample size=100. While sample size=3000, the performance of the 3 methods converge.

Figure 3 shows that for the online shopping dataset, ZTree has overall the best performance. The performance of CART catches up with ZTree at sample size=1000 and sample size=3000. SVM has poor performance at sample size=3000.

Figure 4 shows that for the bike sharing dataset, CART and ZTree have in general similar performance. CART is slightly better at sample sizes 1000 and 3000 and ZTree is slightly better at sample size 1000. SVM does not perform well.

Figure 5 shows that for the online news popularity dataset, SVM has the best performance and ZTree is a close second. CART clearly overfitted the training data when sample size=100, performed worse than a one node model that always predicted the mean of the training data. In fact, ZTree returns one node model 70% of the time when sample size=100.

Several general patterns emerged from this empirical evaluation. 1) Comparing the two decision tree algorithms, ZTree in general has better performance when sample sizes are small, and CART catches up when sample size=1000 or sample size=3000. ZTree ranked the best in four of the five datasets when sample size=100. 2) Compared to the decision tree based methods, SVM performs better in the "adult" dataset but worse in the "shopping" and "bike" datasets. 3) Compared to CART, ZTree grows smaller trees in general. 4) Except for the news popularity dataset, the variations of the tree depth are smaller for ZTree at each sample size setting, which may suggest that the tree structures are more stable.

Table 1

| Dataset | Cases | Z-Tree | CART-sklearn | SVM-sklearn |
|---|---|---|---|---|
| adult | 100 | **0.770** | 0.731 | 0.740 |
| adult | 300 | 0.813 | 0.814 | **0.836** |
| adult | 1000 | 0.853 | 0.850 | **0.879** |
| adult | 3000 | 0.878 | 0.881 | **0.893** |
| banking | 100 | **0.763** | 0.712 | 0.740 |
| banking | 300 | 0.830 | 0.774 | **0.838** |
| banking | 1000 | **0.895** | 0.865 | 0.892 |
| banking | 3000 | **0.918** | 0.909 | 0.915 |
| shopping | 100 | **0.796** | 0.751 | 0.780 |
| shopping | 300 | **0.841** | 0.822 | **0.841** |
| shopping | 1000 | **0.880** | 0.876 | 0.862 |
| shopping | 3000 | **0.906** | 0.905 | 0.869 |
| bike | 100 | **148.0** | 149.1 | 149.3 |
| bike | 300 | **126.0** | **126.0** | 136.4 |
| bike | 1000 | 95.9 | **94.2** | 124.7 |
| bike | 3000 | 72.3 | **71.2** | 114.0 |
| news | 100 | 0.411 | 0.452 | **0.405** |
| news | 300 | 0.404 | 0.413 | **0.399** |
| news | 1000 | 0.395 | 0.397 | **0.394** |
| news | 3000 | 0.390 | 0.390 | **0.386** |

Figure 1

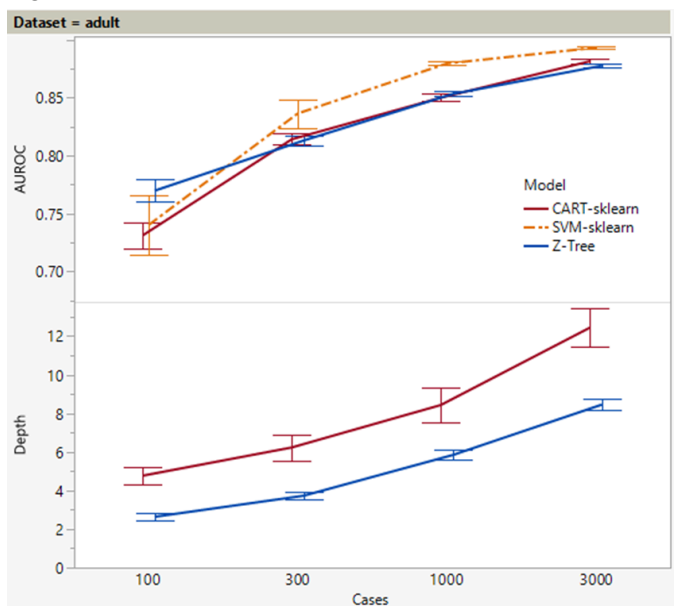

Figure 2

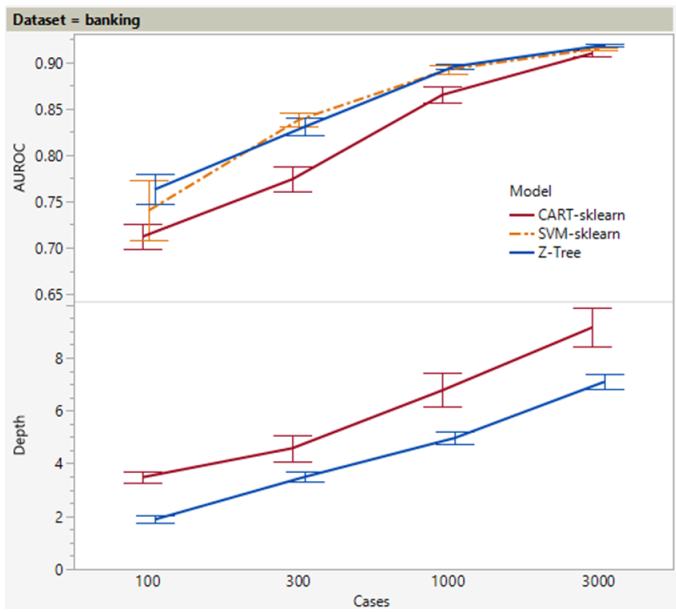

Figure 3

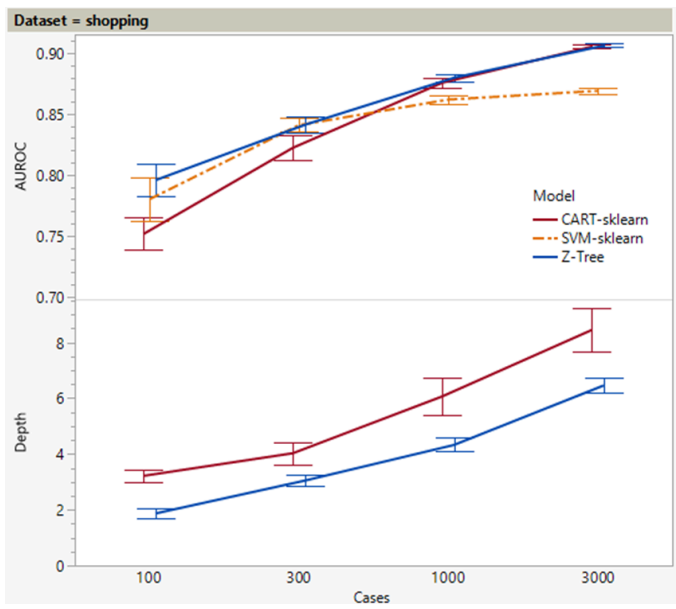

Figure 4

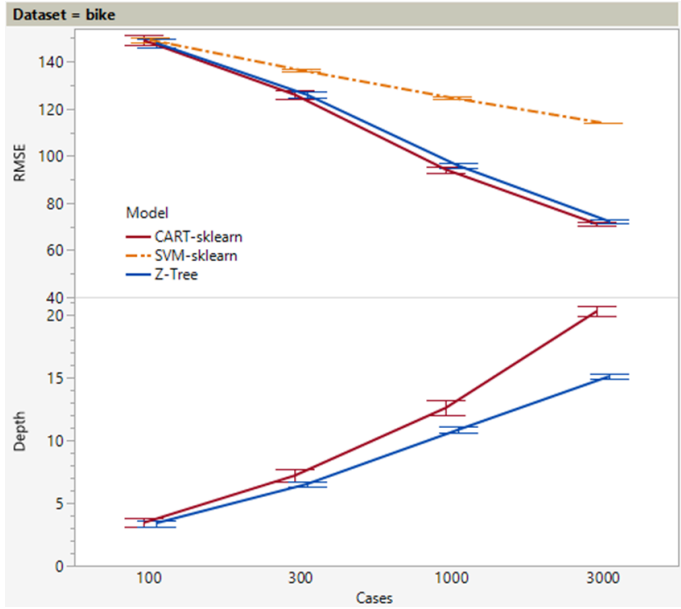

Figure 5

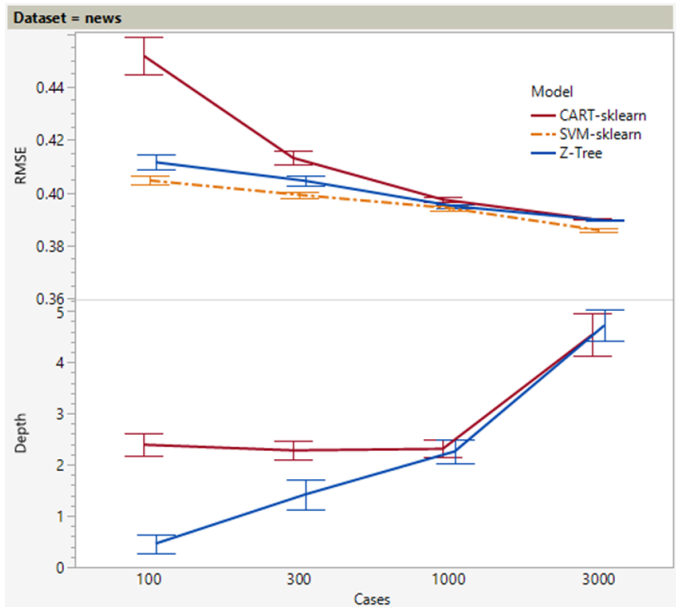

# Discussion and Future Work

As an alternative to node-purity based decision tree learning algorithms, we proposed a statistical subgroup test based decision tree learning approach and demonstrated its strong performance in the empirical comparison with CART and SVM. The proposed approach has a unique way of controlling multiple testing by performing cross-validation to derive adjusted test scores, serving as the splitting vs. stopping criteria. Since the trees are grown in such a controlled manner, no post pruning is needed. The proposed approach has the following advantages:

1) The threshold for the adjusted test score is the parameter for controlling tree complexity and overfitting. This threshold has a clear statistical meaning and can be mapped to a multiple testing adjusted p-value. Different statistical tests can be plugged in the proposed tree learning framework easily.
2) Because of no post-pruning, once a decision tree is trained with a certain threshold, all decision trees with larger threshold are also determined by simply removing the branches that do not meet the larger threshold. This property greatly reduced the burden of parameter optimization. Users can easily examine various tree structures from simple to complex and compare their validation performance without much computational cost, providing a transparent trade-off between complexity and performance.
3) Experimental results show that the proposed method consistently achieves strong predictive performance across both classification and regression tasks, typically exceeding CART when sample size is small. It also tends to create smaller and more stable tree structures, which contributes to model interpretability and reproducibility.

Regarding the learning efficiency, although the CV based splitting vs. stopping criteria is more computational expensive than the standard node purity based score calculation, it eliminates the need of post pruning and makes model tuning highly efficient and intuitive. The overall learning efficiency of ZTree training (including model tuning) is comparable to that of CART. For example, the running time of training an optimal ZTree for the Adult income dataset with 10,000 cases is about 1 minute on an average Intel i5 laptop (external CV for threshold tuning: 10 fold 1 time; internal CV for calculating z score for each node: 5 fold 10 times).

A natural extension of our framework is its application to datasets with treatment, outcome, and covariates, the typical structure used in causal inference and personalized treatment effect modeling. SIDES and GUIDE are other tree based methods for such applications. Simulation data were used to evaluate ZTree's performance in such applications with very encouraging results. Besides precision medicine, we are very interested in applying ZTree to other fields such as marketing strategy and website optimization (A/B testing) etc.

Another promising direction is the development of ensemble versions of our method. By using a larger p-value threshold, our framework can generate weaker learners, which are ideal for ensemble strategies such as bagging, boosting, or random forests. These ensemble techniques could improve predictive performance by reducing bias and variance.